\documentclass[10pt,conference,a4paper]{IEEEtran}
\IEEEoverridecommandlockouts

\usepackage{framed,multirow}
\usepackage{cite}
\usepackage{amsmath,amssymb,amsfonts}
\usepackage{algorithmic}
\usepackage{graphicx}
\usepackage{textcomp}
\usepackage{xcolor}
\usepackage{latexsym}
\usepackage{epstopdf}
\usepackage{multicol}
\usepackage{epsfig}
\usepackage{booktabs}
\usepackage{subfigure}
\usepackage{array}

\def\BibTeX{{\rm B\kern-.05em{\sc i\kern-.025em b}\kern-.08em
    T\kern-.1667em\lower.7ex\hbox{E}\kern-.125emX}}
\begin{document}

\renewcommand{\thefootnote}{\fnsymbol{footnote}}
\title{Semantic Bilinear Pooling for Fine-Grained Recognition}

\author{\IEEEauthorblockN{ Xinjie Li, Chun Yang, Song-Lu Chen, Chao Zhu*\thanks{*Corresponding author}, Xu-Cheng Yin}
\IEEEauthorblockA{\textit{School of Computer and Communication Engineering} \\
\textit{University of Science and Technology Beijing}\\
Beijing, China \\
Email: abcdvzz@hotmail.com, \{ych.learning, chenslvs7\} @gmail.com, \{chaozhu, xuchengyin\} @ustb.edu.cn}
}

\maketitle

\begin{abstract}
Naturally, fine-grained recognition, e.g., vehicle identification or bird classification, has specific hierarchical labels, where fine categories are always harder to be classified than coarse categories. However, most of the recent deep learning based methods neglect the semantic structure of fine-grained objects and do not take advantage of the traditional fine-grained recognition techniques (e.g. coarse-to-fine classification). In this paper, we propose a novel framework with a two-branch network (coarse branch and fine branch), i.e., semantic bilinear pooling, for fine-grained recognition with a hierarchical label tree. This framework can adaptively learn the semantic information from the hierarchical levels. Specifically, we design a generalized cross-entropy loss for the training of the proposed framework to fully exploit the semantic priors via considering the relevance between adjacent levels and enlarge the distance between samples of different coarse classes. Furthermore, our method leverages only the fine branch when testing so that it adds no overhead to the testing time. Experimental results show that our proposed method achieves state-of-the-art performance on four public datasets.
\end{abstract}


\section{Introduction}
\label{sec1}

The fine-grained recognition task focuses on distinguishing sub-classes of the same basic classes, e.g., classification of bird species, or recognition of vehicle models. The main challenges of fine-grained recognition are the subtle inter-class differences and large intra-class diversity. Different from other generic classification tasks, such as text recognition and ImageNet classification \cite{DengDSLL009}, fine-grained recognition task requires more discriminative features. For instance, different vehicles may look similar. However, even for the same car, the images vary a lot due to different poses, various viewpoints, different car upholstery and changing illumination.

\begin{figure}[tb]
\centering{\includegraphics[width=8.75cm]{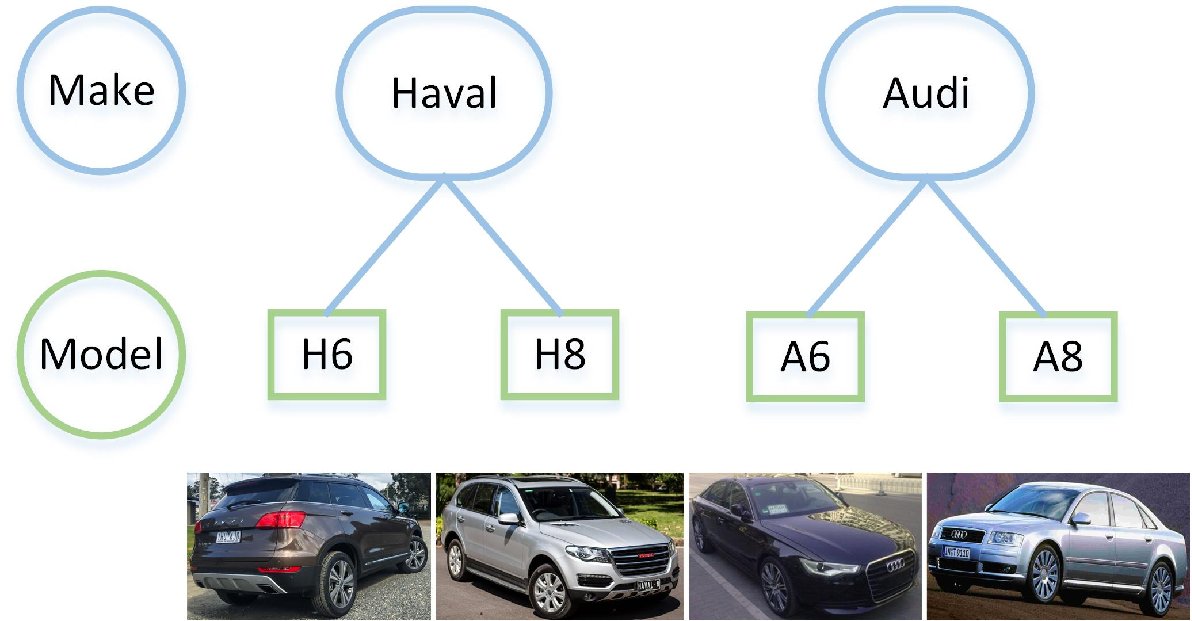}}
\caption{Illustration of the semantic structure. The samples used in this figure are from the Haval series and the Audi series, where from top to bottom are make and model.}
\label{fig1}
\end{figure}

\begin{figure*}[tb]
\centering{\includegraphics[width=17.4cm]{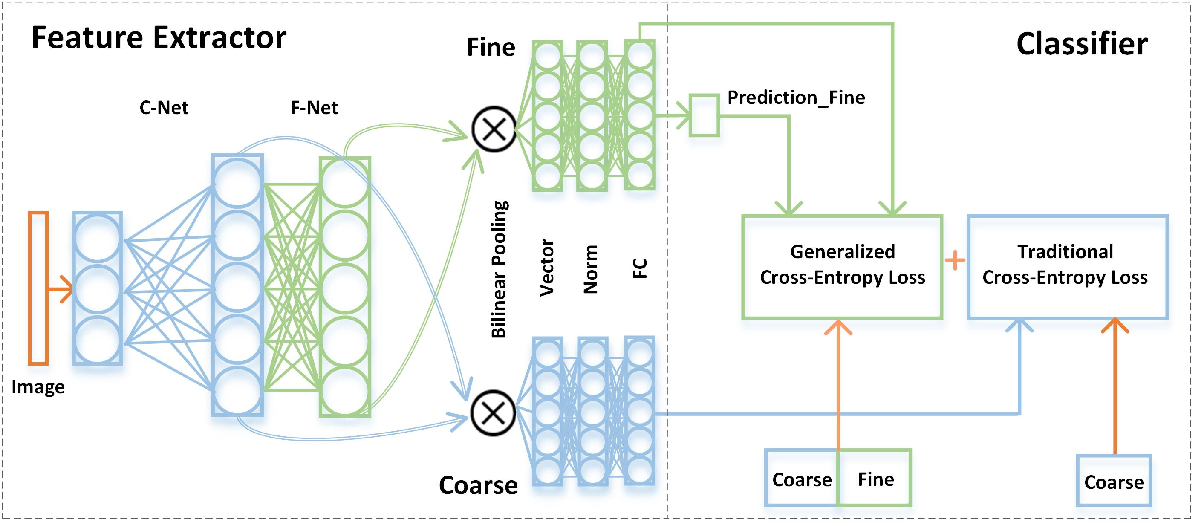}}
\caption{SBP-CNN structure. Blue lines indicate how coarse branch flows and green lines represent the fine branch. The coarse branch comes after C-Net (coarse network) while the fine branch comes after F-Net (fine network). After the bilinear vectors, the norm (short for normalization) part includes a signed square root layer and an L2 normalization layer. FC (fully connected layer) produces final feature distribution for classification.}
\label{fig5}
\end{figure*}

A lot of methods have been proposed for fine-grained recognition with good performance. For example, FCAN \cite{LiuXWL16} achieves $89.1\%$ accuracy and BCNN \cite{LinRM15} achieves $91.3\%$ accuracy on the Stanford Cars dataset \cite{KrauseStarkDengFei-Fei_3DRR2013}, etc. However, most of them only focus on how to localize discriminative regions and represent subtle visual differences. Unlike other generic objects, fine-grained datasets often have a unique tree structure as shown in Fig. \ref{fig1} which shows the label relations of CompCars dataset \cite{YangLLT15}: make and model. Although a large number of work studies \cite{LuZ19, ZhouL16, ChenWGDLL18} on hierarchical multi-label learning, they typically use the traditional basic CNN models rather than applying the fine-grained methods.

In this paper, we propose a novel framework with hierarchical label tree which includes two main contributions: (1) We propose a new deep framework with a two-branch network, i.e., semantic bilinear pooling, by incorporating the bilinear pooling \cite{LinRM15} method with the semantic structure of objects, as shown in Fig. \ref{fig5}. (2) We design a novel loss to fully exploit the priors so that the results of the coarse branch can guide the predictions of the fine branch.

The rest of the paper is organized as follows. Section 2 covers the related work on state-of-the-art fine-grained recognition methods. Section 3 introduces the details of our proposed method. Section 4 gives the experimental evaluation that we conducted. Section 5 concludes the whole paper and brings up some discussions for future work.

\section{Related Work}
As mentioned previously, two main difficulties of fine-grained classification are how to localize discriminative regions and represent subtle visual differences. To tackle these problems, plenty of methods have been raised these years. FCAN \cite{LiuXWL16} used a fully convolutional attention localization network based on reinforcement learning which uses an attention module to locate multiple parts simultaneously. MAMC \cite{SunYZD18} combined metric learning with visual attention regions. NTS \cite{YangLWHGW18} detected attention regions in a reinforcement-learning manner. SWP \cite{HuWLS17} used attention masks to guide the pooling operation. RACNN \cite{FuZM17} proposed a recurrent attention mechanism to learn subtle features on different scales. And MACNN \cite{FuZM17} adopted a channel grouping module to generate different attention maps. In recent literature, MGE \cite{learning} utilized a mixture of neural networks to learn diverse distribution. S3N \cite{selective} used class peak responses to localize informative regions. Luo et al. \cite{cross} proposed Cross-X learning to learn multi-scale features. Chen et al. \cite{ChenBZM19} proposed DCL to learn robust fine-grained features by solving jigsaw puzzles.

Meanwhile, a bilinear structure(BCNN) \cite{LinRM15} also attracted people. Lin et al. \cite{LinRM15} applied two CNN streams as two feature extractors and multiplied their outputs using the outer product at each discriminative part. Furthermore, they added matrix power normalization in  \cite{LinM17}. However, a problem of the original bilinear pooling lies in its high dimension, thus some methods have been then proposed to solve this problem. Kong et al. proposed LRBP \cite{kong2017lowrankbilinear} to reduce high feature dimensionality with kernelized modules. Gao et al. proposed CBP \cite{GaoBZD16} using Tensor Sketch projection and Random Maclaurin projection to largely reduce dimension without sacrificing too much accuracy. Besides, kernel pooling \cite{CuiZWLLB17} applied the Gaussian RBF kernel to catch higher-order feature interactions. Cai et al. \cite{CaiZZ17} proposed a polynomial kernel-based model to capture higher-order statistics. G$^2$DeNet \cite{WangLZ17} utilized a global Gaussian distribution embedding to pool discriminable features. MoNet \cite{GouXCS18} combined the G$^2$DeNet and bilinear pooling. iSQRT-COV \cite{LiXWG18} applied Newton-Schulz iteration to the training process to get a better performance on GPU. However, these approaches did not consider embedding label relations in their work.

Few methods are exploring the semantic label relations. BGL \cite{ZhouL16} leveraged the label hierarchy using bipartite-graph labeling but it was complex to optimize. CLC \cite{LuZ19} incorporated coarse labels into a sigmoid cross- entropy function but it used multi-label learning which did not fully exploit label hierarchy. HSE \cite{ChenWGDLL18} applied probability embedding and label regularization but it used four-hierarchy information and more convolution layers. And these methods did not use bilinear pooling operation. Compared to these methods, our method uses a two-branch network and generalized cross-entropy loss function to fully explore the semantic relations. Furthermore, our method uses less information and easy to implement.

\section{Proposed Method}
In this section, we introduce our semantic bilinear pooling convolutional neural network (SBP-CNN) from two aspects: one is our two-branch network and the other is our generalized cross-entropy loss function.

\subsection{Two-Branch Network}
Given an input image $x$, we first extract coarse-branch image feature maps ${f_I} \in \mathbb{R}^{D^1 \times H^1 \times W^1}$
($D^1$, $H^1$ and $W^1$ denote the channel number, height and width of the coarse-level feature maps) by
\begin{equation}
\label{coarse}
  f_I = f(x)
\end{equation}
where $f(\cdot)$ is a coarse-branch feature extractor that is implemented by a network(e.g., the C-Net in Fig. \ref{fig5}). Then we merge the width dimension and height dimension of ${f_I} \in \mathbb{R}^{D^1 \times H^1 \times W^1}$ to get ${f_{IM}} \in \mathbb{R}^{D^1 \times H^1W^1}$. Afterwards, we apply bilinear pooling method \cite{LinRM15} (i.e., outer product) to $f_{IM}$:
\begin{equation}
\label{product}
 F = \frac{1}{H W}{{f}_{IM}} {{f}_{IM}}^{\mathrm{T}}
\end{equation}
where $F$ is a Gram matrix representing second-order statistics of the image. Next, we vectorize and normalize $F$ as  \cite{LinRM15} proposed to get the final coarse-branch representation:
\begin{equation}
\label{representation}
 F_R = 
 \frac{{\operatorname{sign}( {\operatorname{vec} F} )\sqrt{\left|{\operatorname{vec} F}\right|}}}{\|{\operatorname{sign}( {\operatorname{vec} F} )\sqrt{\left|{\operatorname{vec} F}\right|}}\|}
\end{equation}
Finally, we get the coarse-branch feature distribution $z_f$ for classification:
\begin{equation}
\label{distribution}
 z_f = \psi\left(F_R\right)
\end{equation}
where $\psi(\cdot)$ is implemented by a fully connected layer.

For the fine-branch feature maps $g_I$, we extract them by
\begin{equation}
\label{fine}
  g_I = g(f_I) = g(f(x))
\end{equation}
where $g(\cdot)$ is a fine-branch feature extractor that is implemented by a network(e.g., the F-Net in Fig. \ref{fig5}). Then we perform same operations to $g_I$ to get $z_g$ for fine-branch classification. Note that we only use fine branch in Fig. \ref{fig5} when testing.

Compared to traditional one-branch classification, two-branch classification increases the representation power of the network by adding another constraint. Furthermore, because the coarse branch has semantic relations with the fine branch, coarse-branch representation could be regarded as prior information to fine-branch representation. Intuitively, coarse-branch classification aims to find coarse information such as shape and size while fine-branch classification tends to focus on more detailed information like headlights of cars and heads of birds. As shown in Fig. \ref{fig1}, cars of Haval series are different from cars of Audi series in shape while cars of Audi A6 differ from cars of Audi A8 in headlights.

\begin{figure}[tb]
\centering{\includegraphics[width=8.75cm]{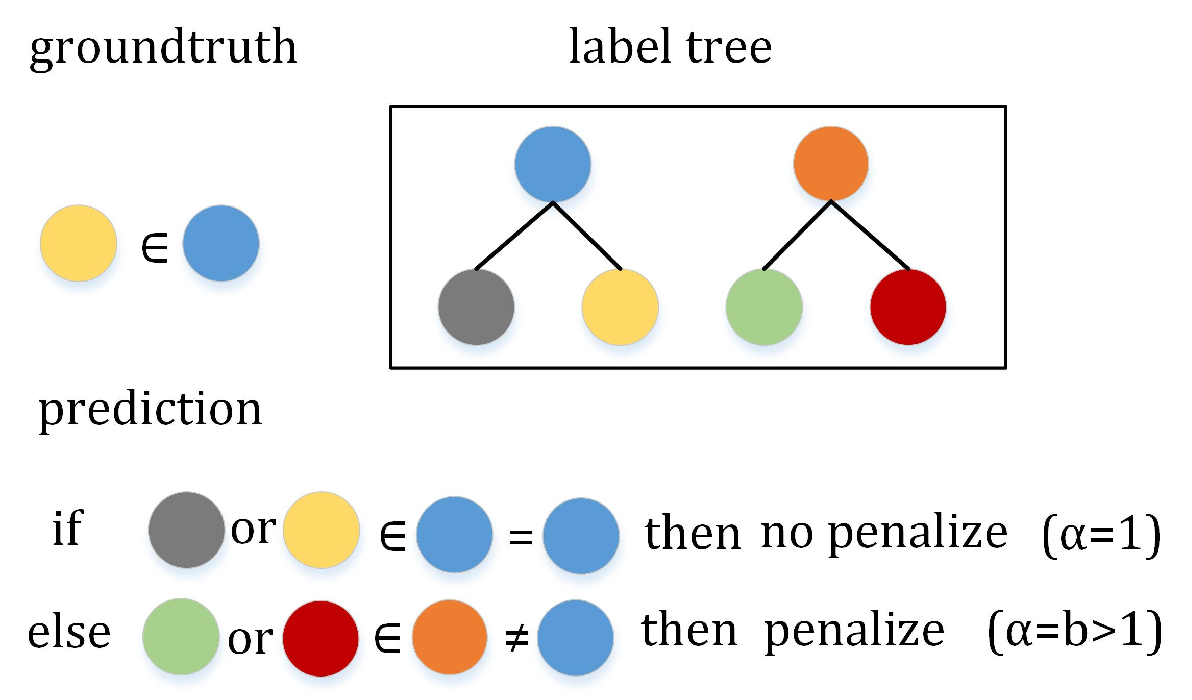}}
\caption{Illustration for choosing the value of $\alpha$. Here, $b$ is the hyperparameter which indicates the value of the penalty term.}
\label{fig66}
\end{figure}

\begin{figure}[tb]
\centering{\includegraphics[width=4.375cm]{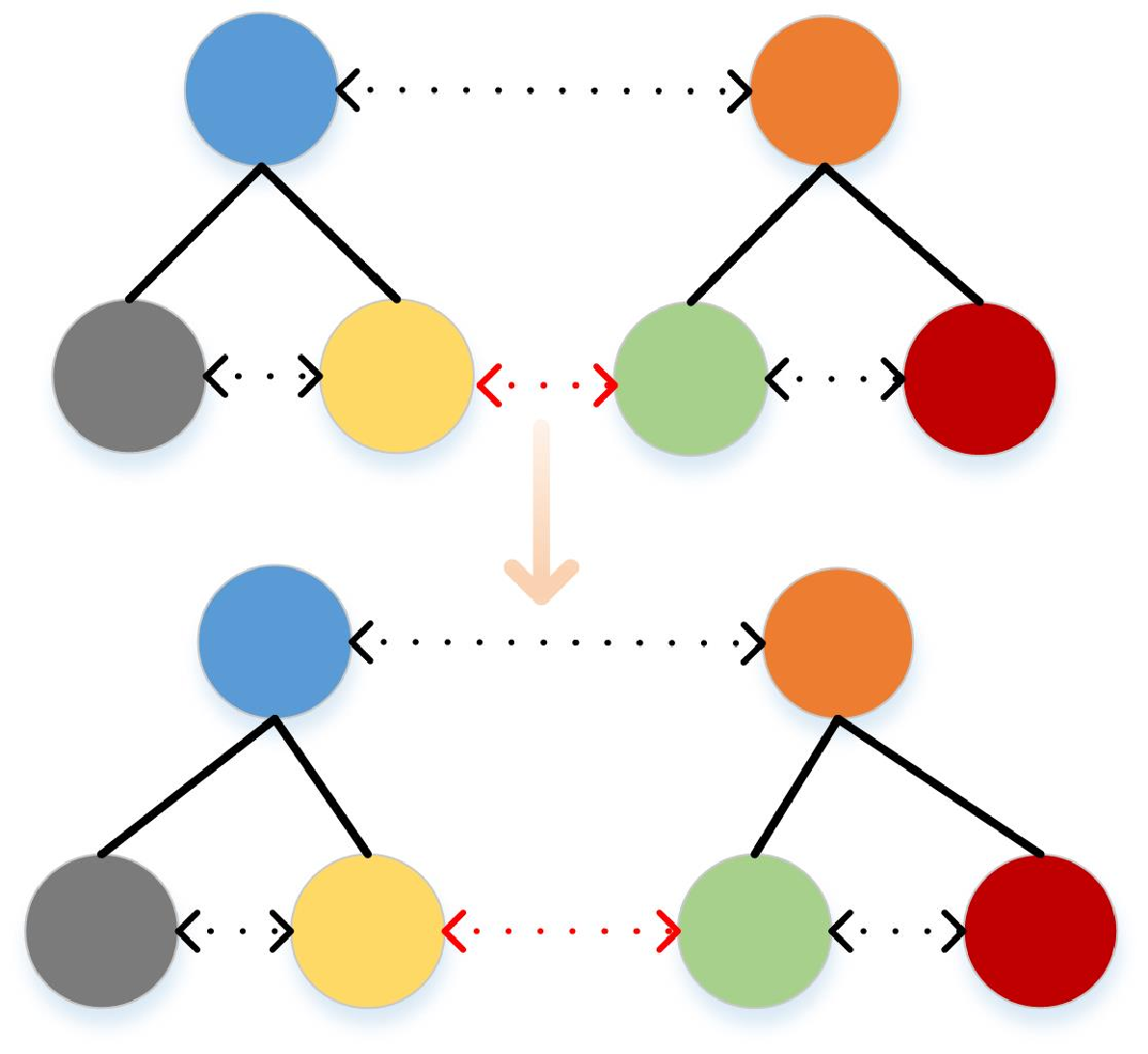}}
\caption{Illustration for the implicit meaning of generalized cross-entropy loss function. Note that the distance between the yellow circle and the green circle becomes larger.}
\label{fig88}
\end{figure}

\subsection{Generalized Cross-Entropy Loss Function}

The traditional cross-entropy loss function takes all distances between predictions and corresponding labels as equal. This assumption is not suitable for our SBP-CNN because it is more harmful to the representation capacity of the model that predictions do not match the basic label relations between two levels (e.g., for the Audi-Audi A8 category, Audi-Audi A6 prediction is more acceptable than Audi-Haval H3 prediction). So we proposed our generalized cross-entropy loss function which is defined in Eq.(\ref{loss}):
\begin{equation}
\label{loss}
  Loss=-\sum_{i=1}^{n}\sum_{c=1}^{C}\alpha_iy_{ic}\log(a_{ic})
\end{equation}
where $n$ is the number of batch size and $i$ denotes the
\begin{math}
  i^{th}
\end{math} sample in this batch. $C$ represents the total number of categories. Normally, \begin{math}
  y_i
\end{math} equals to 1 when $i^{th}$ sample belongs to category $c$, otherwise it equals to 0. And \begin{math}
  a_c
\end{math} is defined in Eq.(\ref{alpha}):
\begin{equation}\label{alpha}
  a_{c}=\frac{e^{z_c}}{\sum_{k=1}^{C}e^{z_k}}
\end{equation}
where $z$ is a vector, which has $C$ dimensions, produced by fully connected layers of the network. And
$\alpha_i$ is a penalty term. It can be observed that when $\alpha_i = 1$ the loss function degenerates into the traditional cross-entropy loss function. Normally, its value will equal to $b$ ($> 1$) when predictions do not match the label relations between two levels.

We take advantage of the paired labels to make the decision whether we penalize this sample or not. The paired labels are defined as $[c^1_i, c^2_i]$, where $c^1_i$ and $c^{2}_i$ denote the coarse category and fine category of the $i^{th}$ sample, respectively. Besides, other fine categories which  do not belong to $c^1_i$ is denoted by $c^{other}_i$. Furthermore, the decision making procedure is defined as:
\begin{equation}
\alpha_i=\left\{\begin{array}{ll}{b>1,} & {if\ {p}^2_{i} \in c^{other}_i } \\ {1,} & {  else }\end{array}\right.
\end{equation}
where ${p}^2_{i}$ denotes the prediction of fine level. An explicit explanation is shown in Fig. \ref{fig66}. In this way, we are intended to fully exploit the prior knowledge so that this kind of semantic regulation can guide the training process. From another perspective, distance among fine categories which do not belong to the same coarse category should be larger as shown in Fig. \ref{fig88}. 

Note that we use the traditional cross-entropy loss function in the coarse branch because one fine label corresponds to only one coarse label. Furthermore, the final loss function is a combination of generalized cross-entropy loss function of the fine branch and traditional cross-entropy loss function of the coarse branch. The loss weight ratio 
is defined as:
\begin{equation}
r=W_t:W_g
\end{equation}
where $W_t$ denotes loss weight of traditional cross-entropy loss function and $W_g$ denotes loss weight of generalized cross-entropy loss function. We set $r> {1:1}$ because coarse-branch representation is the base of fine-branch representation as shown in Eq.\ref{fine}.

\section{Experiments}
\subsection{Datasets}

\begin{table}[tb]
\caption{Data distribution of the datasets.}
\begin{center}
\begin{tabular}{|c|c|c|c|c|}
\hline
    Datasets&$\#$Coarse&$\#$Fine&$\#$Train&$\#$Val\\
    \hline
    CompCars \cite{YangLLT15}&75&431&16016&14939\\
    StanfordCars \cite{KrauseStarkDengFei-Fei_3DRR2013}&49&196&8144&8041\\
    CUBbirds \cite{WahCUB_200_2011}&70&200&5994&5794\\
    Aircrafts \cite{maji13fine-grained}&70&100&6667&3333\\
  \hline
\end{tabular}
\label{tab5}
\end{center}
\end{table}

We conducted experiments on four benchmarks including the Stanford Cars dataset \cite{KrauseStarkDengFei-Fei_3DRR2013}, CompCars dataset \cite{YangLLT15}, CUBbirds \cite{WahCUB_200_2011} and Aircraft \cite{maji13fine-grained}.

According to the corresponding fine labels provided officially, we construct four label trees for four datasets separately. For example, 'Audi A8' is provided officially and we split it into 'Audi' and 'Audi A8' as our coarse label and fine label, respectively. In this way, we construct 75 make labels and 431 model labels for CompCars dataset \cite{YangLLT15}, 49 make labels and 196 model labels for Stanford Cars dataset \cite{KrauseStarkDengFei-Fei_3DRR2013}, 70 family labels and 200 species labels for CUBbirds dataset \cite{WahCUB_200_2011}, 70 family labels and 100 variant labels for Aircraft dataset \cite{maji13fine-grained}.

There are semantic hierarchical relations between adjacent levels and exclusion relations among the same level. Detailed data distribution of four datasets is listed in Table \ref{tab5}.

\begin{table}[tb]
\caption{Different $r$ and corresponding accuracy on CUBbirds \cite{WahCUB_200_2011} dataset using CBP-based SBP-CNN.}
\begin{center}
 \begin{tabular}{|c|c|c|c|c|c|}
 \hline
  $r$ & $1:1$ & $3:2$ & $7:3$ & $4:1$ & $9:1$\\
    \hline
  Accuracy(\%) & 83.3 & 84.6 & 84.8 & 84.0 & 82.9\\
  \hline
\end{tabular}
\label{tab0}
\end{center}
\end{table}

\begin{figure}[tb]
\centering{\includegraphics[width=8.75cm]{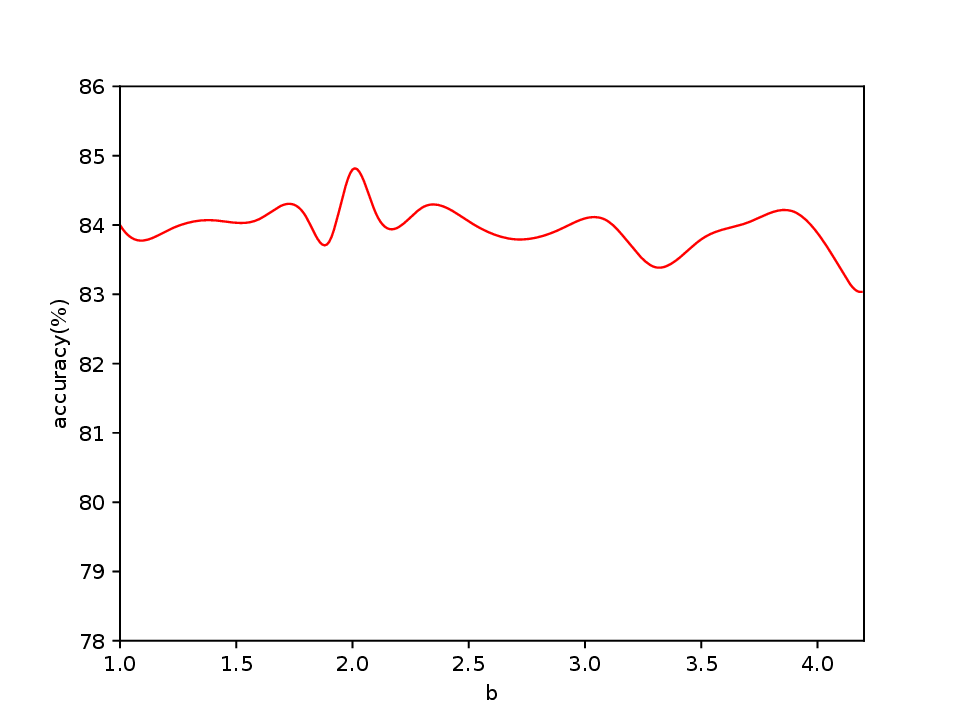}}
\caption{Changing accuracy($\%$) of the validation dataset of CUBbirds \cite{WahCUB_200_2011} using CBP-based SBP-CNN. Here, $b$ is the hyperparameter which indicated the value of the penalty term. Intuitively, when $b$ equals to 2.0, accuracy reaches the vertex.}
\label{fig8}
\end{figure}

\subsection{Implementation Details}
\subsubsection{Baselines}
We use the CBP \cite{GaoBZD16} method and the iSQRT-COV \cite{LiXWG18} method as our baselines. Compared to original BCNN \cite{LinRM15}, they have lower dimension and faster convergence rate. Furthermore, in CBP-based experiments, we use VGG16 \cite{SimonyanZ14a} as the base network and we use VGG16 \cite{SimonyanZ14a} or ResNet50 \cite{HeZRS16} as the base network in iSQRT-COV-based experiments.

\subsubsection{Experiments on SBP-CNN}
In all our experiments, we crop one image into ten patches with a size of 448x448 as the input images. And we choose SGD as our optimization method with momentum in 0.9 during the training process. We implement C-Net with $conv1\_1-conv4\_3$ and F-Net with $conv5\_1-conv5\_3$ in VGG16-based experiments. And we implement C-Net with earlier 41 convolutional layers of ResNet50 and F-Net with following 9 convolutional layers in ResNet50-based experiments. We set $b=2$ and $r=7:3$ through extensive experiments as shown in Fig. \ref{fig8} and Table \ref{tab0}. We perform all experiments using Caffe \cite{JiaSDKLGGD14} or PyTorch \cite{NEURIPS2019_9015} over two NVIDIA TITAN Xp GPUs.

For experiments based on the CBP \cite{GaoBZD16} method, we take two steps to train the network as CBP does. Firstly, we finetune the classifier in Fig. \ref{fig5} without training the feature extractor. Secondly, we train the entire network. In the first step, we set the initial learning rate to $1.0$ and it decays by a factor of 10 for every 30 epochs. And we finetune the model from ImageNet pretrained model for 100 epochs with weight decay of $5\times10^{-6}$. In the second step, we set the learning rate to $0.001$ and fix it. And we finetune the model from the first-step model for 30 epochs with weight decay of $5\times10^{-4}$.

For experiments based on the iSQRT-COV \cite{LiXWG18} method, we train the model from ImageNet pretrained model for 100 epochs with weight decay of $0.001$ in an end-to-end manner. We set the initial learning rate to $0.0012$ for feature extractor and $0.006$ for the classifier. And the learning rate decays by a factor of 10 for every 30 epochs.

\subsubsection{Ablation Analysis}

\begin{table*}[tb]
\caption{Comparison of the accuracy($\%$) of baselines, our method that only uses Generalized Cross-Entropy loss(GCE) or only utilizes Two-Branch network(TB), where the method with $*$ indicates that we implemented the experiments by our own. And 'w/o' indicates without.}
\begin{center}
\begin{tabular}{|c|c|c|c|c|c|c|c|}
\hline
     Backbone&Method &  TB & GCE  & CompCars & StanfordCars & Birds & Aircrafts \\
    \hline
 \multirow{8}{*}{{VGG16}}
 &CBP \cite{GaoBZD16}&$-$&$-$&$*$94.0&$*$90.8&84.0&$*$87.4\\
    &Ours w/o GCE & $\checkmark$ &&94.3&91.3&84.3&88.2\\
    &Ours w/o TB && $\checkmark$& 94.7 & 91.6 & 84.5 & 88.9 \\
   &\textbf{Ours(CBP)} &$\checkmark$&$\checkmark$&\textbf{95.2}&\textbf{91.9}&\textbf{84.8}&\textbf{89.3}\\
    \cline{2-8}
    &iSQRT-COV \cite{LiXWG18} &$-$&$-$&$*$96.3&92.5&87.2&90.0\\
    &Ours w/o GCE &$\checkmark$&&96.7&92.9&87.4&90.6\\
    &Ours w/o TB && $\checkmark$ & 96.8 & 92.9 & 87.5 & 90.8 \\
    &\textbf{Ours(iSQRT-COV)}& $\checkmark$ & $\checkmark$&\textbf{97.0}&\textbf{93.2}&\textbf{87.8}&\textbf{91.1}\\
    \hline
 \multirow{4}{*}{{ResNet50}} &iSQRT-COV \cite{LiXWG18} &$-$ &$-$&$*$96.9&92.8&88.1&90.0\\
    &Ours w/o GCE & $\checkmark$&&97.3&93.5&88.5&90.5\\
    & Ours w/o TB&  & $\checkmark$ &97.4&93.7&88.3&91.2\\
    &\textbf{Ours(iSQRT-COV)}  & $\checkmark$ & $\checkmark$&\textbf{97.8}&\textbf{94.3} &\textbf{88.9}&\textbf{91.7}\\
    \hline
\end{tabular}
\label{ablation}
\end{center}
\end{table*}

\begin{figure}[tb]
\centering{\includegraphics[width=8.75cm]{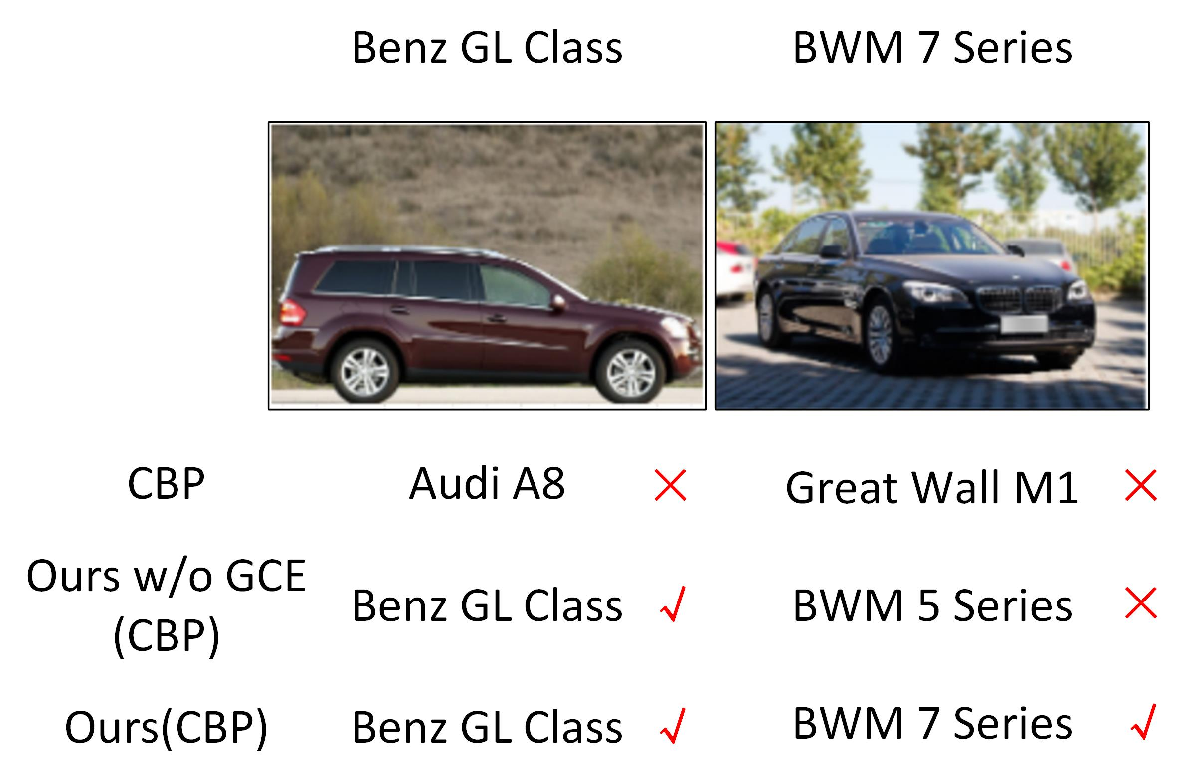}}
\caption{Two test samples from CompCars dataset \cite{YangLLT15} and their predictions. GCE is Generalized Cross-Entropy loss.}
\label{fig18}
\end{figure}

\begin{figure}[tb]
\centering{\includegraphics[width=8.75cm]{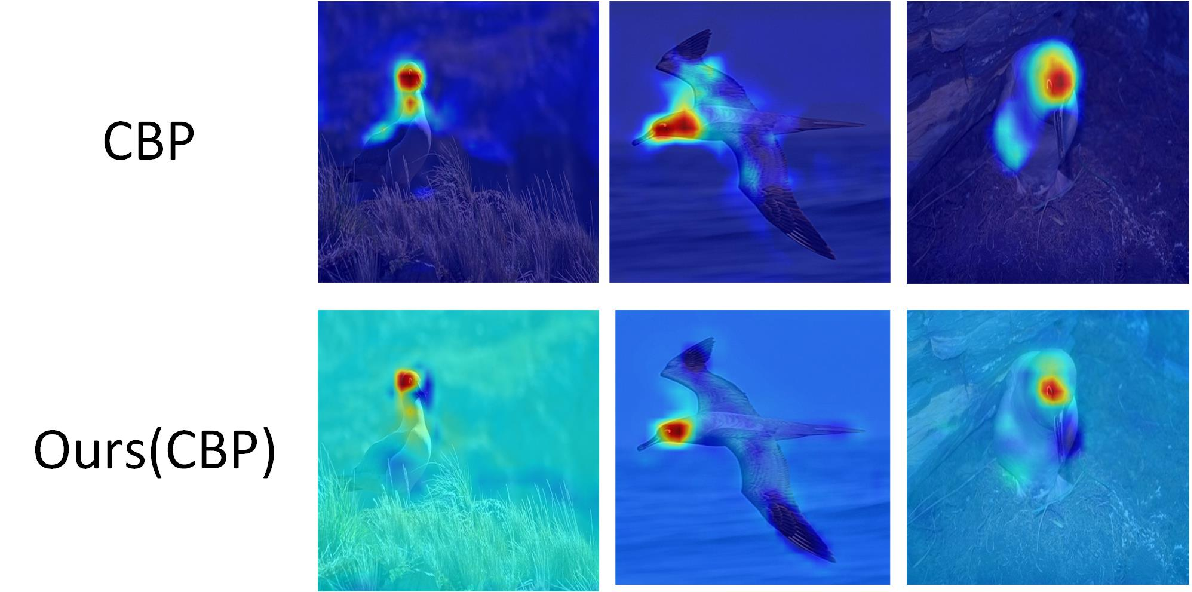}}
\caption{Visualization of the attention map. The samples are from Sooty$\_$Albatross class of validation dataset of CUBbirds \cite{WahCUB_200_2011}.}
\label{heatmap}
\end{figure}

To fully investigate our method, we provide a detailed ablation analysis on different settings of variants as shown in Table \ref{ablation}. Intuitively, both the two-branch network and the generalized cross-entropy loss function provide a better performance, but generalized cross-entropy loss function offers more performance improvement in most cases. We further illustrate this phenomenon in Fig. \ref{fig18}. It is worth noting that our method provides performance improvement without sacrificing speed, adding parameters or dimensions as we only use fine branch when testing which is the same with baselines. Furthermore, we visualize the attention regions detected by CBP-based SBP-CNN as shown in Fig. \ref{heatmap}. It can be observed that our model focuses on the more discriminative region than CBP \cite{GaoBZD16} does and ours is more robust to different pose and background.

\subsection{Results}
\subsubsection{Comparison with Semantic Methods}

\begin{table}[tb]
\caption{Comparison of accuracy ($\%$) of our method and other semantic methods on two benchmarks. 'S-cars' represents StanfordCars for better visualization.}
\begin{center}
\begin{tabular}{|c|c|c|c|}
\hline
     Backbone&Method & S-Cars & Birds\\
    \hline
 \multirow{3}{*}{{{VGG16}}}
 &BGL \cite{ZhouL16}& 86.0&75.9\\
   &\textbf{Ours(CBP)} &91.9&84.8\\
    &\textbf{Ours(iSQRT-COV)}&\textbf{93.2}&\textbf{87.8}\\
    \hline
 \multirow{3}{*}{{ResNet50}}
  &CLC \cite{LuZ19}&$-$&79.3\\
   &HSE \cite{ChenWGDLL18}&$-$&88.1\\
    &\textbf{Ours(iSQRT-COV)}  &\textbf{94.3} &\textbf{88.9}\\
    \hline
\end{tabular}
\label{semantic}
\end{center}
\end{table}

We compare our method with other methods that utilize extra coarse information. As shown in Table \ref{semantic}, our method achieves the best performance on two benchmarks (other methods did not report results on the other two datasets). Note that our method significantly outperforms BGL \cite{ZhouL16} and CLC \cite{LuZ19}. Furthermore, HSE \cite{ChenWGDLL18} utilizes four-level information and adds more convolution layers while we only use two-level information and do not add any convolution layers. And our method still outperforms HSE by 0.7$\%$. 

\subsubsection{Comparison with State-of-the-Art Methods}

In Table \ref{sota}, we compare our method with most of the fine-grained visual classification methods using the VGG16 network or the ResNet50 network. Our method based on iSQRT-COV \cite{LiXWG18} achieves the best performance on all four benchmarks with remarkable margins.

\begin{table*}[tb]
\caption{Comparison of accuracy ($\%$) on four benchmarks.}
\begin{center}
\begin{tabular}{|c|c|c|c|c|c|}
\hline
     Backbone&Method & CompCars & StanfordCars & Birds & Aircrafts \\
    \hline
 \multirow{13}{*}{{VGG16}}
    &FCAN \cite{LiuXWL16}&$-$&89.1&82.0&$-$\\
    &LRBP \cite{kong2017lowrankbilinear}&$-$&90.9&84.2&87.3\\
    &KP \cite{CuiZWLLB17}&$-$&92.4&86.2&86.9\\
    &iBCNN \cite{LinM17}&$-$&92.0&85.8&88.5\\
    &G$^2$DeNet  \cite{WangLZ17}&$-$&92.5&87.1&89.0\\
    &HIHCA \cite{CaiZZ17}&$-$&91.7&85.3&88.3\\
    &MoNet \cite{GouXCS18}&$-$&90.8&85.7&88.1\\
    &SWP \cite{HuWLS17}&95.3&90.7&$-$&$-$\\
    &BCNN \cite{LinRM15}&$*$93.0&90.6&84.0&86.9\\
    &CBP \cite{GaoBZD16}&$*$94.0&$*$90.8&84.0&$*$87.4\\
    &iSQRT-COV \cite{LiXWG18} &$*$96.3&92.5&87.2&90.0\\
   \cline{2-6}
    &\textbf{Ours(CBP)} &95.2&91.9&84.8&89.3\\
    &\textbf{Ours(iSQRT-COV)}&\textbf{97.0}&\textbf{93.2}&\textbf{87.8}&\textbf{91.1}\\
   \hline
    \multirow{2}{*}{{{VGG19}}}
    &RACNN \cite{FuZM17}&$-$&92.5&85.3&88.2\\
    &MACNN \cite{ZhengFML17}&$-$&92.8&86.5&89.9\\
    \hline
 \multirow{7}{*}{{ResNet50}} 
 &SWP \cite{HuWLS17}&97.5&92.3&$-$&$-$\\
 &NTS \cite{YangLWHGW18}&$-$&93.9&87.5&91.4\\
 &MAMC \cite{SunYZD18}&$-$&93.0&86.5&$-$\\
 &DFL \cite{WangMD18}&$-$&93.1&87.4&91.7\\
 &KP \cite{CuiZWLLB17}&$-$&91.9&84.7&85.7\\ 
 &iSQRT-COV \cite{LiXWG18} &$*$96.9&92.8&88.1&90.0\\
 \cline{2-6}
    &\textbf{Ours(iSQRT-COV)}&\textbf{97.8}&\textbf{94.3}&\textbf{88.9}&\textbf{91.7}\\
\hline
\end{tabular}
\label{sota}
\end{center}
\end{table*}

\section{Conclusion and Future work}
In this paper, we proposed a novel fine-grained recognition method named Semantic Bilinear Pooling, which incorporates the hierarchical label tree and bilinear pooling together with a two-branch network. In this way, we can use semantic connections between different levels with the bilinear pooling method and they will reinforce each other during training. Moreover, we generalized the traditional cross-entropy loss function to the generalized one which aims to fully exploit the priors and enlarge the distance between samples of different coarse classes. Experiments showed that our method is effective for the fine-grained classification task. 

In the future, we will further study the proposed SBP-CNN in two directions, i.e., how to effectively incorporate hierarchical label tree with other methods, and how to apply different forms of label relations like a graph which contains more information for reasoning.

\section*{Acknowledgment}

This work was partly supported by National Natural Science Foundation of China (61703039 and 62072032), Beijing Natural Science Foundation (4194084 and 4174095) and Fundamental Research Funds for the Central Universities (FRF-TP-18-060A1).

\bibliographystyle{IEEEtran}
\bibliography{egbib}

\end{document}